\documentclass[a4paper, 10pt, conference]{ieeeconf}

\IEEEoverridecommandlockouts                              

\usepackage{amsmath,amssymb}
\usepackage{graphicx}
\usepackage{booktabs}
\usepackage{multirow}
\usepackage{xcolor}
\usepackage{url}
\usepackage[hidelinks]{hyperref}
\usepackage{cleveref}
\usepackage{comment}
\usepackage{subcaption}

\begin{document}

\title{\LARGE \bf
Few-Shot Adaptation to Non-Stationary Environments via Latent Trend Embedding for Robotics
}

\author{%
  Yasuyuki Fujii$^{1}$,\;Emika Kameda$^{1}$,\;Hiroki Fukada$^{2}$,\;Yoshiki Mori$^{3}$ \;Tadashi Matsuo$^{4}$, and\;Nobutaka Shimada$^{1}$%
  \thanks{$^{1}$College of Information Science and Engineering, Ritsumeikan University, Osaka, Japan.{\ttfamily fujiyasu@fcritsumei.ac.jp}}%
  \thanks{$^{2}$Production \& Technology Department, NIPPN CORPORATION, Tokyo, Japan.}%
  \thanks{$^{3}$University of Osaka, Osaka, Japan.}%
  \thanks{$^{4}$National Institute of Technology, Ichinoseki College, Iwate, Japan.}%
}

\bstctlcite{IEEEexample:BSTcontrol}
\maketitle
\begin{abstract}
Robotic systems operating in real-world environments often suffer from concept shift, where the input–output relationship changes due to latent environmental factors that are not directly observable.
Conventional adaptation methods update model parameters, which may cause catastrophic forgetting and incur high computational cost.

This paper proposes a latent Trend ID–based framework for few-shot adaptation in non-stationary environments.
Instead of modifying model weights, a low-dimensional environmental state,
referred to as the Trend ID, is estimated via backpropagation while the model parameters remain fixed.
To prevent overfitting caused by per-sample latent variables, we introduce temporal regularization and a state transition model that enforces smooth evolution of the latent space.

Experiments on a quantitative food grasping task demonstrate
that the learned Trend IDs are distributed across distinct
regions of the latent space with temporally consistent
trajectories, and that few-shot adaptation to unseen
environments is achieved without modifying model parameters.

The proposed framework provides a scalable and interpretable solution
for robotics applications operating across diverse and evolving environments.
\end{abstract}

\section{Introduction}
Automating robots in real-world environments remains a challenging problem due to dynamic environmental changes and uncertainty in the controlled system~\cite{Lesort2020}. Among these challenges, concept shift—where the properties of the controlled system change with environmental variation—is a particularly critical issue. Concept shift refers to the phenomenon in which the relationship between inputs and outputs varies due to latent environmental factors, even though the distribution of observed inputs remains unchanged~\cite{Gama2014,Lu2019}. For example, the moisture content and density of food materials fluctuate over time and with changes in temperature and humidity. Since such variations cannot be observed through visual sensors, the same food item may exhibit different grasped weights despite having an identical appearance~\cite{Fukada2023}. To address this type of environmental variation, machine learning-based adaptation methods have attracted considerable attention in recent years~\cite{Bayram2022,liu2024task}. However, because concept shift significantly degrades the performance of pre-trained models, there is a growing need for learning frameworks capable of real-time adaptation in non-stationary environments.

Conventional approaches to concept shift have relied on transfer learning~\cite{Pan2010} and meta-learning~\cite{Finn2017}. These methods, however, overwrite model parameters during adaptation to new environments, and therefore carry the risk of catastrophic forgetting—the loss of previously acquired knowledge~\cite{Kirkpatrick2017,VanDeVen2024,josifovski2024continual}. Furthermore, in operational settings where the environment changes frequently, the computational cost of retraining the model each time it does so imposes a significant practical constraint.
\begin{figure}[tb]
    \centering
    \includegraphics[width=\columnwidth]{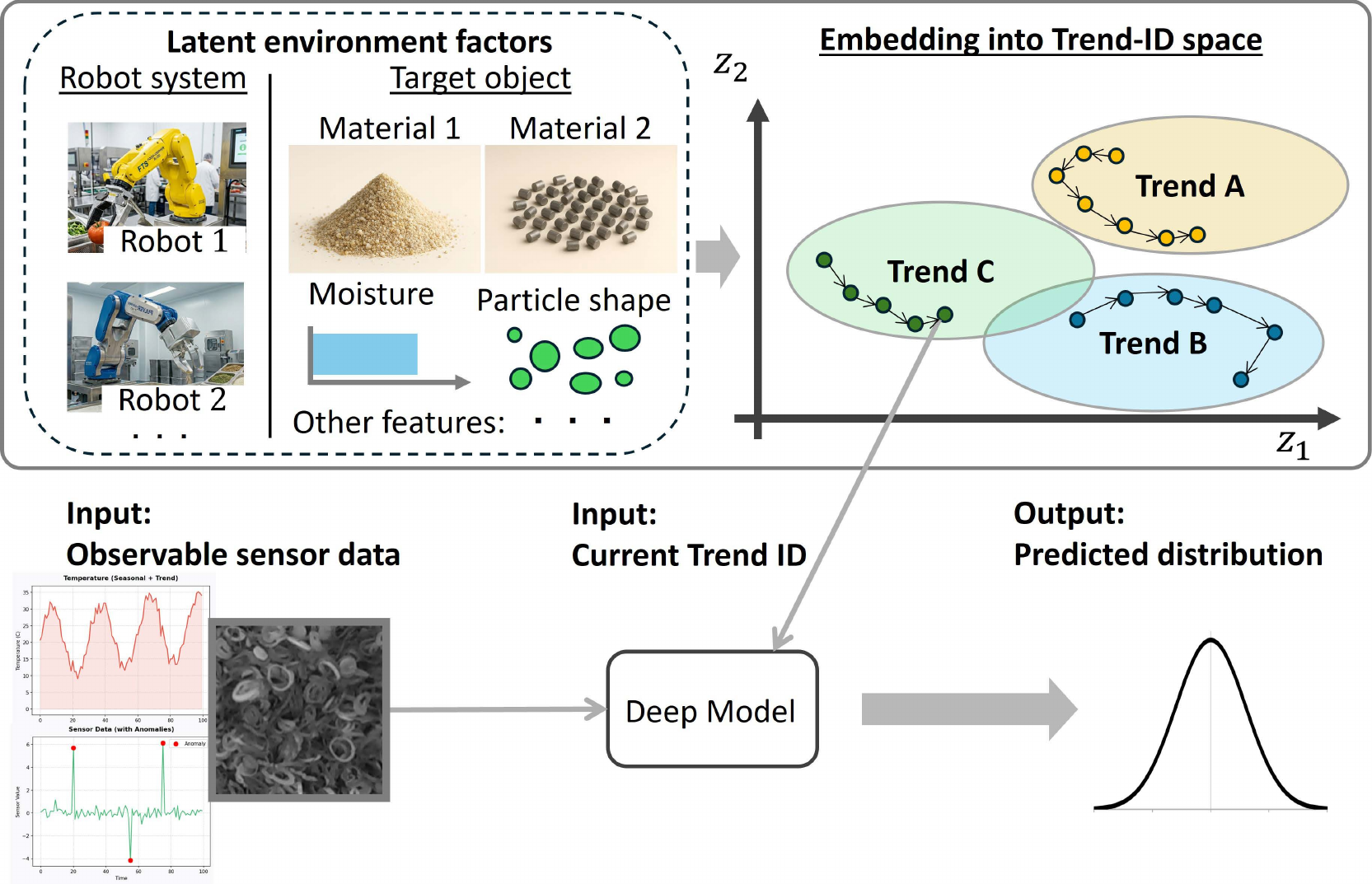}
    \caption{Overview of the proposed Trend ID framework. Latent environment factors—such as robot configuration, target material properties, and ambient conditions—are embedded as low-dimensional Trend IDs in a continuous latent space. The deep model receives both observable sensor data and the estimated Trend ID to output a predicted distribution adapted to the current environment.}
    \label{fig:concept}
\end{figure}

In contrast to conventional approaches that update model parameters to accommodate environmental changes, the proposed method fixes the model and instead provides a variable representing the environmental state as an additional input. Specifically, we define the \textit{trend} as the hidden environmental state that governs system behavior but does not appear explicitly in the observed data, and represent it as a low-dimensional vector in a latent space, referred to as the \textit{Trend ID}. At test time, with the model weights held fixed, only the Trend ID is estimated from a small amount of data via backpropagation~\cite{aoyama2024few}. As illustrated in Fig.~\ref{fig:concept}, this formulation enables rapid adaptation to new environments while retaining knowledge of all environments within a single Trend space. In particular, for environments exhibiting \textit{recurrence}—where similar environmental states reappear over time—the continuous latent space of Trend IDs enables interpolation and retrieval of past experience, facilitating prediction in previously unseen environments. However, assigning a unique Trend ID to each training sample inherently introduces the risk of \textit{ID leak}, in which the model relies solely on the Trend ID while ignoring input features, potentially leading to severe degradation in generalization performance. To mitigate this overfitting risk, we combine temporal constraints based on a state transition model with regularization terms, thereby encouraging the model to appropriately leverage both input features and trend information.

The proposed framework is applicable to a broad range of tasks subject to concept shift. It is particularly well-suited for settings in which large amounts of data are accumulated across diverse objects and conditions—such as high-mix low-volume production lines in franchise operations or multi-site robotic systems—where the key practical advantage lies in the ability to sequentially adapt to new environmental conditions without discarding previously acquired knowledge. To validate its effectiveness, this paper focuses on a quantitative grasping task for granular food materials~\cite{kadokawa2023learning,niu2023goats}. In this task, the moisture content and density of the food material vary in a manner unobservable by visual sensors, giving rise to a prototypical instance of concept shift. Following the approach of Fukada et al.~\cite{Fukada2023}, we construct a model that estimates grasped weight as a probability distribution rather than a point estimate, and apply the proposed framework on top of this model to achieve grasp control that adapts to environmental variation while accounting for uncertainty.

The main contributions of this work are threefold:
\begin{enumerate}
    \item \textbf{A framework that avoids catastrophic forgetting}:
    By dynamically controlling Trend IDs for objects with similar characteristics, data from diverse environments and target objects can be processed in a unified manner within a single latent space. Moreover, since model parameters are never overwritten, the framework enables adaptation to new environments while preserving previously acquired knowledge~\cite{wan2024lotus,daab2024incremental}.

    \item \textbf{Rapid adaptation via few-shot inference}:
    By restricting updates to the Trend ID alone and suppressing overfitting through temporal constraints, the framework achieves convergence to the current environmental state from a small number of observations within a short time~\cite{aoyama2024few,vecerik2024robotap}.

    \item \textbf{Interpretability in the latent space}:
    Since estimated Trend IDs are represented as vectors in a latent space, the environmental states across different time steps and food materials can be quantitatively compared and visualized~\cite{pore2024dear}. This facilitates both interpretation of model behavior and analysis of environmental variation.
\end{enumerate}

\section{Related Work}

\subsection{Catastrophic Forgetting and Continual Learning}
In methods that sequentially update model parameters, catastrophic forgetting—the loss of previously acquired knowledge upon adaptation to new data—poses a fundamental challenge~\cite{Kirkpatrick2017}. Continual learning addresses this issue through three main families of approaches: regularization-based methods~\cite{Kirkpatrick2017,josifovski2024continual}, replay-based methods~\cite{Rolnick2019}, and dynamic architecture methods~\cite{Rusu2016,wan2024lotus}~\cite{VanDeVen2024}. Nevertheless, all of these approaches face increasing computational and memory demands as the number of tasks grows~\cite{Lesort2020}, making frequent retraining impractical in settings where environmental conditions vary continuously~\cite{josifovski2024continual}.

\subsection{Adaptation via Conditioned Inputs}
A complementary line of research has focused on controlling model behavior through conditioned inputs while keeping model parameters fixed. \cite{Hausman2018} demonstrated that feeding a task-representing latent variable into a policy enables a single model to execute multiple skills. PEARL~\cite{Rakelly2019} and CNP~\cite{Garnelo2018} further extended this idea by inferring latent variables from a small number of experiences or context points to condition predictions. More recently, disentangled representations of environment and agent state have been shown to improve generalization under distribution shift~\cite{pore2024dear}, and implicit latent representations inferred from observation history have demonstrated robust performance under domain shift in locomotion tasks~\cite{wang2024toward}.

In the domain of robot manipulation, Diffusion Policy~\cite{Chi2023} proposed a framework that generates action sequences conditioned on observation history, achieving strong performance across a range of tasks. Context-conditioned approaches have also been applied to deformable object manipulation~\cite{thach2024defgoalnet} and to bridging the sim-to-real gap through online adaptation of a parameter subset while keeping the base policy fixed~\cite{zhang2024bridging}. A systematic survey of conditioning strategies for integrating foundation models into robotics has also been presented in~\cite{Firoozi2025}.

\subsection{Positioning of the Proposed Method}
The proposed method builds upon these conditioned input approaches while differing from existing methods in two fundamental respects.

First, whereas PEARL and CNP infer latent variables implicitly from experience or context, the Trend ID in the proposed method is an explicit environmental state vector estimated directly from observed data via backpropagation~\cite{pore2024dear,wang2024toward}. This transparency enables visualization and quantitative comparison of estimated environmental states, thereby improving the interpretability of the model.

Second, whereas the skill embeddings of~\cite{Hausman2018} condition the model on \textit{what to do} (i.e., the type of task), the Trend ID conditions the model on \textit{under what environmental conditions} (i.e., the state of physical conditions). This distinction allows the proposed framework to simultaneously achieve adaptation to continuous concept shift and avoidance of catastrophic forgetting, without updating any model parameters~\cite{daab2024incremental,zhang2024bridging}.

Furthermore, since the Trend ID is designed not to suppress observable input information, the model retains the ability to adapt through visual features even for objects with differing appearances~\cite{aoyama2024few,vecerik2024robotap}. The essential strength of the proposed framework lies in its applicability to settings where large amounts of data are accumulated across diverse objects and conditions—such as multi-site production lines or multi-robot systems—enabling adaptation to unseen environments through interpolation and retrieval of past environmental states, without discarding any previously acquired knowledge.

\section{Concept Shift Adaptation Framework Using Trend IDs}
\label{sec:method}

In this section, we propose an approach to concept shift adaptation based on Trend IDs. Rather than adapting model parameters, we \textbf{adapt a low-dimensional latent representation of the environmental state, referred to as the Trend ID}. This formulation allows a single model to dynamically condition its predictions on changing environmental conditions without any parameter updates.

\subsection{Problem Formulation and Approach}
\begin{figure*}[tb]
    \centering
    \begin{minipage}{0.49\textwidth}
        \centering
        \includegraphics[width=\textwidth]{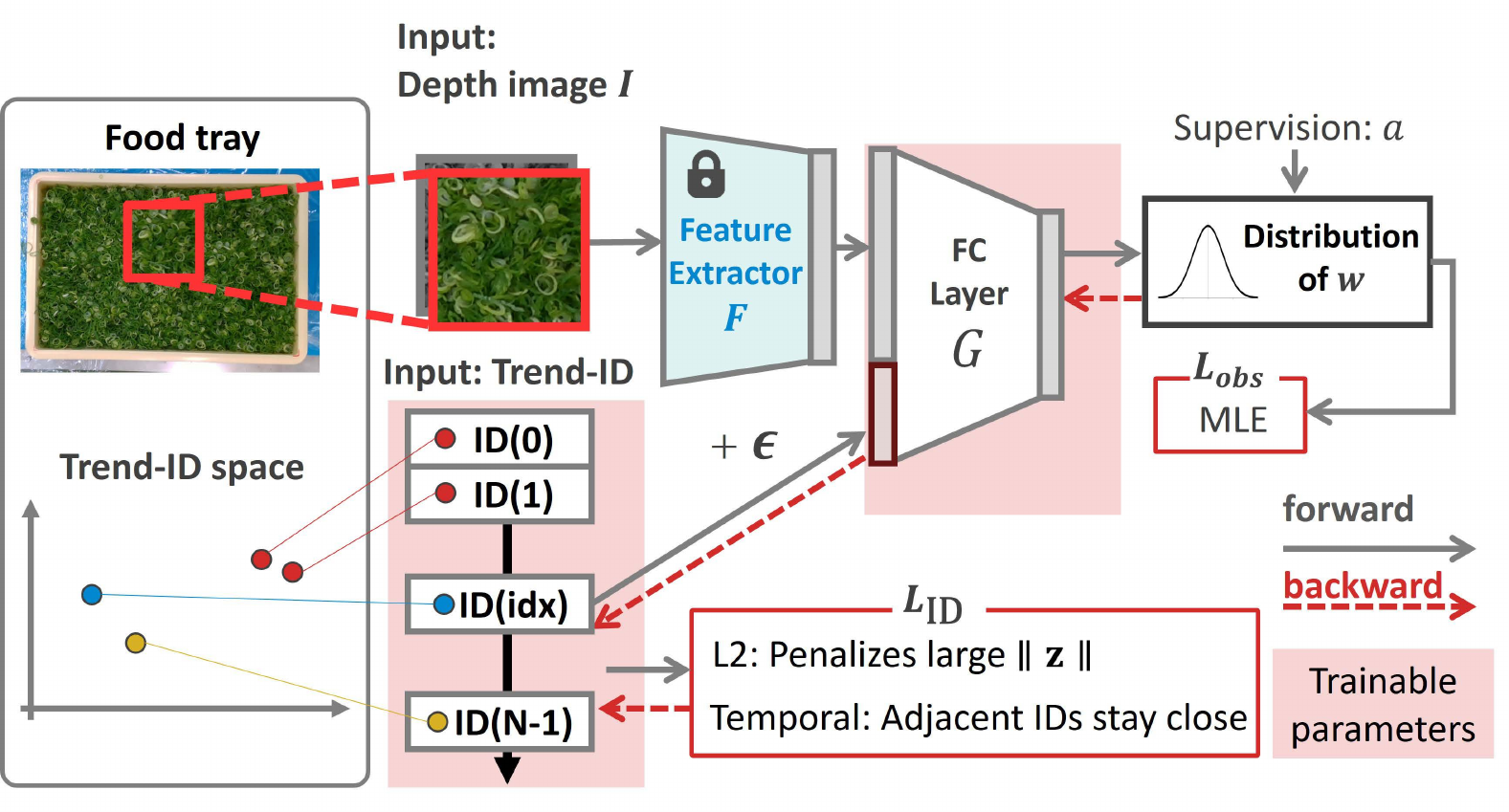}
        \subcaption{Training phase}
        \label{fig:model_train}
    \end{minipage}
    \hfill
    \begin{minipage}{0.49\textwidth}
        \centering
        \includegraphics[width=\textwidth]{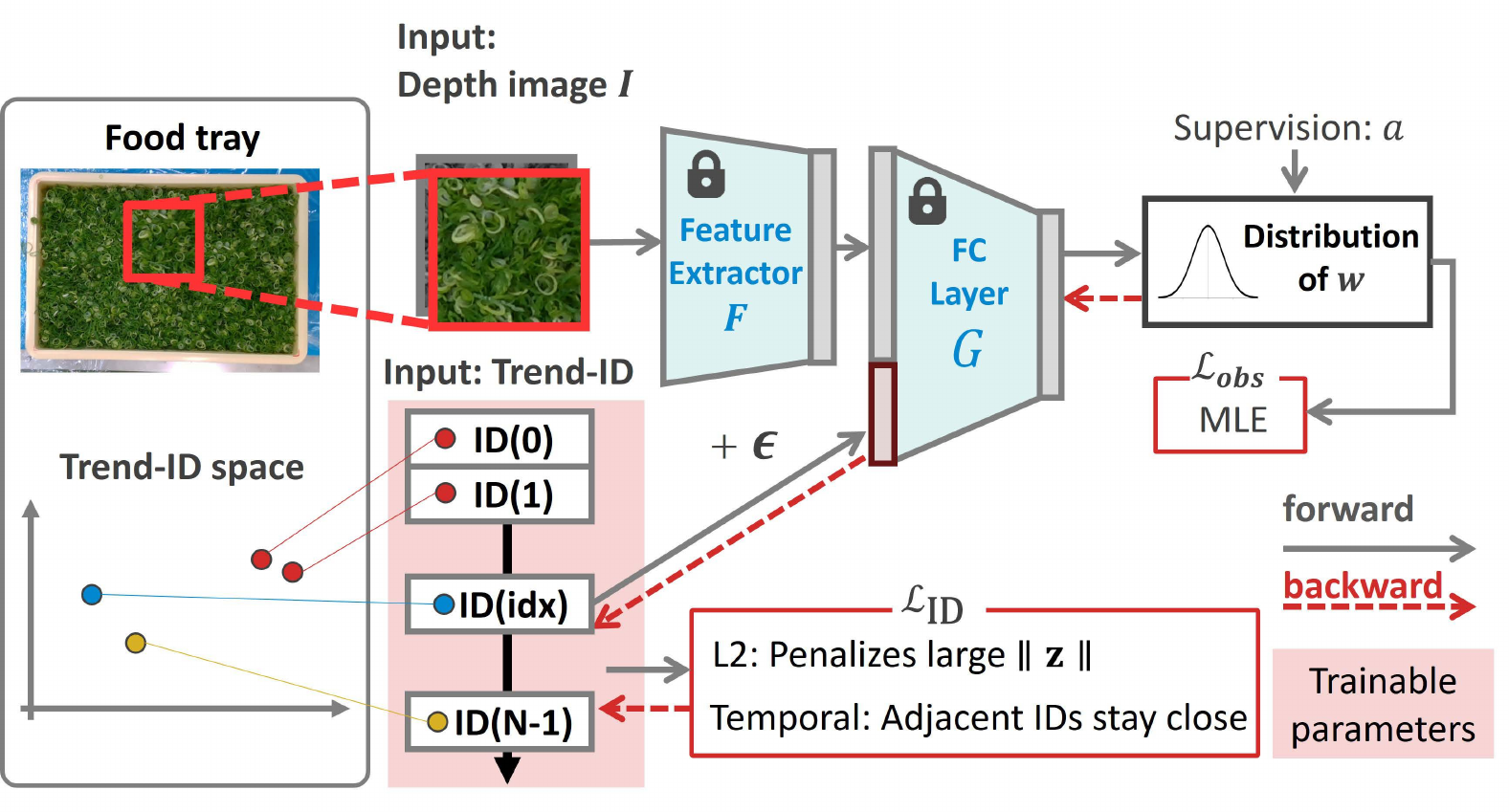}
        \subcaption{Test phase}
        \label{fig:model_test}
    \end{minipage}
    \caption{Architecture of the proposed Trend-ID based adaptation framework. (a) During training, both the fully connected layer $G$ and the Trend IDs $\{\mathbf{z}_i\}$ are jointly optimized while the feature extractor $F$ is held fixed. (b) During testing, both $F$ and $G$ are frozen, and only the Trend ID $\mathbf{z}^{\text{test}}$ is optimized from a small number of observed samples via backpropagation.}
    \label{fig:model}
\end{figure*}

\subsubsection{Concept Shift with Recurrence}
Let $x_t \in \mathcal{X}$ denote the observed input at time $t$, $a_t \in \mathcal{A}$ the controllable action, and $y_t \in \mathcal{Y}$ the observed outcome. Notably, even for the same $x_t$, the outcome $y_t$ varies substantially depending on the choice of $a_t$, and executing $a_t$ irreversibly alters $x_t$, making re-trials fundamentally infeasible. Under stationary conditions, the relationship $y_t \sim P(Y \mid x_t, a_t; \boldsymbol{\Theta})$ holds. In the presence of concept shift, however, the input–output relationship is modulated by a latent environmental state $\mathbf{z}_t \in \mathbb{R}^d$:
\begin{equation}
    y_t \sim P(Y \mid x_t, a_t, \mathbf{z}_t; \boldsymbol{\Theta})
\end{equation}
where $\mathbf{z}_t$ cannot be directly observed and must be inferred from data.

\subsubsection{Trend ID as a Latent Environmental State}
We instantiate $\mathbf{z}_t$ in Eq.~(1) as the \textbf{Trend ID} $\mathbf{z}_t \in \mathbb{R}^d$. Unlike discrete domain labels, the Trend ID is embedded in a continuous latent space where proximity reflects similarity in environmental conditions. This continuous representation enables generalization across similar conditions, interpolation between observed environmental states, and quantitative comparison of environmental conditions. The Trend ID serves simultaneously as a model input that conditions predictions and as a learnable parameter optimized via backpropagation during both training and inference.

\subsubsection{Inference Strategy: Few-Shot Trend Estimation}
During training, a learnable Trend ID $\mathbf{z}_i$ is assigned to each training sample $(x_i, a_i, y_i)$ and jointly optimized with the model to construct a structured latent space. As illustrated in Fig.~\ref{fig:model_train}, the Trend IDs and the fully connected layer $G$ are jointly updated via backpropagation, while the feature extractor $F$ remains fixed.

At test time, upon encountering a new environment, the model parameters $\boldsymbol{\Theta}$ are held fixed, and only the Trend ID $\mathbf{z}_{\text{test}}$ is optimized via gradient descent to minimize the prediction error over a small set of observed samples $\{(x_j^{\text{test}}, a_j^{\text{test}}, y_j^{\text{test}})\}_{j=1}^{M}$ ($M = 5$--$10$), as shown in Fig.~\ref{fig:model_test}. Using the estimated $\mathbf{z}_{\text{test}}$ for subsequent predictions within the same environment enables rapid few-shot adaptation while fully preserving previously acquired knowledge.

\subsubsection{Overfitting Risk and Regularization in Trend ID Learning}
The design of assigning a unique learnable Trend ID to each training sample inherently carries the risk of overfitting. As training progresses, \textbf{ID leak} may occur, wherein the model relies almost exclusively on the Trend ID rather than the observed input $x_t$ to make predictions. In this degenerate case, the Trend ID effectively acts as a ground-truth label for each sample, severely degrading generalization to new environments.

To address this issue, we impose multiple regularization
constraints on the Trend IDs. Specifically, we combine: (1)
addition of Gaussian noise to the Trend ID as data augmentation
during training to improve robustness; (2) a state transition
loss $\mathcal{L}_{\varepsilon}$ that constrains the temporal
evolution of the Trend ID based on a state transition model;
(3) a velocity consistency loss $\mathcal{L}_{v}$ that
penalizes excessively large positional transitions in the latent
space; and (4) a position consistency loss $\mathcal{L}_{p}$
that promotes directional smoothness of the latent trajectory
by penalizing abrupt changes in the direction of motion.
Together, these regularization terms suppress overfitting while
enabling the Trend space to appropriately capture environmental
variation, as described in Section~\ref{sec:regularization}
and further developed in Section~IV.

\subsection{Base Model Architecture}
This work builds upon the probabilistic regression model proposed in~\cite{Fukada2023}. The problem addressed by this model is one in which the outcome $y_t$ varies substantially depending on the choice of action $a_t$, even for the same observation $x_t$. The model therefore adopts a framework in which a neural network outputs the parameters of the function $y_t = f(x_t \mid a_t; \boldsymbol{\Theta})$. Specifically, the model estimates the parameters of the conditional distribution $\mathcal{N}(\mu_t, \sigma_t^2)$ of the target variable $y_t$ given observation $x_t$ and action $a_t$, and is trained by minimizing the negative log-likelihood of the pointwise observations $y_t$.

As shown in Fig.~\ref{fig:model_train}, the model consists of a feature extractor $F$ and a fully connected layer $G$. $F(x_t)$ extracts a feature vector $\mathbf{f}_t$ from the observation, and $G(\mathbf{f}_t)$ outputs regression parameters $\boldsymbol{\theta}, \boldsymbol{\tau}$ from the feature vector. These parameters determine the probability distribution of the target variable given action $a_t$~\cite{Fukada2023}.

Training is performed by minimizing the negative log-likelihood:
\begin{equation}
  \mathcal{L}_{\text{obs}}
  = \frac{(y_t - \mu_t)^2}{\sigma_t^2} + \log(2\pi\sigma_t^2).
  \label{eq:obs_loss}
\end{equation}

\noindent\textbf{Integration of the Trend ID:}
The input to the fully connected layer $G$ is extended from $\mathbf{f}_t$ alone to the concatenation $[\mathbf{f}_t; \mathbf{z}_t]$, incorporating the Trend ID $\mathbf{z}_t \in \mathbb{R}^d$. The feature extractor $F$ is kept fixed, and only the parameters of $G$ and the Trend IDs are subject to training. This separation ensures that $F$ retains environment-agnostic general-purpose features, while $G$ and $\mathbf{z}_t$ are responsible for environment-specific adaptation. For further details of the derivation, the reader is referred to~\cite{Fukada2023}.

\subsection{Learning Model with Trend IDs}
\subsubsection{Training Phase: Construction of the Trend Space}
A unique Trend ID $\mathbf{z}_i \in \mathbb{R}^d$ is assigned to each training sample $(x_i, a_i, y_i, t_i)$. The Trend ID functions both as a model input that conditions predictions and as a learnable parameter optimized via backpropagation, as depicted in Fig.~\ref{fig:model_train}.

Given $N$ training samples, the trainable parameters are restricted to the following:
\begin{itemize}
  \item The weights of the fully connected layer $G$
  \item The Trend IDs $\{\mathbf{z}_i\}_{i=1}^N$ associated with each sample
\end{itemize}
The input to $G$ is the concatenation $[\mathbf{f}_t; \mathbf{z}_t]$. The feature extractor $F$ is fixed when a suitable pre-trained model is available, as is the case for RGB images where transfer learning is applicable. When no appropriate pre-trained model exists, such as for depth images, $F$ may also be included among the trainable parameters.

The model is trained by minimizing the total loss defined in Eq.~\eqref{eq:total_loss}.
Through this optimization, Trend IDs of samples collected under the same environmental conditions with the same robot and target object are drawn together in the latent space, forming a structured Trend space.

\subsubsection{Test Phase: Few-Shot Trend Estimation}
In a new environment, given a small set of labeled samples $\{(x_j^{\text{test}}, a_j^{\text{test}}, y_j^{\text{test}})\}_{j=1}^{M}$, only the Trend ID $\mathbf{z}^{\text{test}}$ is optimized as illustrated in Fig.~\ref{fig:model_test}:
\begin{equation}
  \mathbf{z}^{\text{test}}
  = \arg\min_{\mathbf{z}}
  \sum_{j=1}^{M}
  \mathcal{L}_{\text{obs}}(y_j^{\text{test}} \mid x_j^{\text{test}}, a_j^{\text{test}}, \mathbf{z}).
\end{equation}
The model parameters $F$ and $G$ are held fixed, fully preserving all previously acquired knowledge. The estimated $\mathbf{z}^{\text{test}}$ is then used to generate predictions for new samples in the same environment.

\subsection{Regularization of Trend IDs}
\label{sec:regularization}
The design of assigning a unique Trend ID to each training sample inherently carries the risk of overfitting through ID leak. This section describes the regularization strategy employed to mitigate this problem.

\subsubsection{Composition of the Regularization Term}
The total training loss is defined as a linear combination of four 
terms:
\begin{equation}
  \mathcal{L} = \alpha \mathcal{L}_{\text{obs}} 
              + \beta \mathcal{L}_{\varepsilon} 
              + \gamma \mathcal{L}_{v} 
              + \zeta \mathcal{L}_{p},
  \label{eq:total_loss}
\end{equation}
where $\alpha$, $\beta$, $\gamma$, and $\zeta$ are hyperparameters 
that control the relative contribution of each term.
$\mathcal{L}_{\text{obs}}$ is the observation loss defined in 
Eq.~\eqref{eq:obs_loss}. The remaining three terms constitute the 
regularization applied to the Trend IDs, as described below.

\subsubsection{State Transition Loss}
The state transition loss $\mathcal{L}_{\varepsilon}$ constrains the
temporal evolution of the Trend ID so that it does not significantly
deviate from a prescribed state transition model. By penalizing large
process noise $\boldsymbol{\varepsilon}_i$, deviations from the
nominal trajectory are permitted only when required to fit the
observed data. The concrete form of $\mathcal{L}_{\varepsilon}$
is derived in Section~IV in conjunction with the state transition
model.

\subsubsection{Velocity Consistency Loss}
The velocity consistency loss $\mathcal{L}_{v}$ penalizes excessively 
large positional transitions between adjacent Trend IDs in the latent 
space:
\begin{equation}
  \mathcal{L}_{v} = \sum_{i=1}^{N-1} 
  \frac{(\mathbf{z}_{i+1} - \mathbf{z}_i)^{\top}
        (\mathbf{z}_{i+1} - \mathbf{z}_i)}
       {\sigma_{v}^2 \,\Delta t_i},
  \label{eq:loss_v}
\end{equation}
where $\sigma_{v}^2$ is a hyperparameter reflecting the characteristic 
scale of positional change. This term prevents the Trend ID from 
making excessively large jumps in the latent space between consecutive 
samples.

\subsubsection{Position Consistency Loss}
The position consistency loss $\mathcal{L}_{p}$ promotes smoothness 
of the latent trajectory by penalizing abrupt changes in the direction 
of motion:
\begin{equation}
  \mathcal{L}_{p} = \sum_{i=1}^{N-1} 
  \left\{1 - \frac{\dot{\mathbf{z}}_{i+1}}{\|\dot{\mathbf{z}}_{i+1}\|}
             \cdot
             \frac{\dot{\mathbf{z}}_{i}}{\|\dot{\mathbf{z}}_{i}\|}
  \right\},
  \label{eq:loss_p}
\end{equation}
where $\dot{\mathbf{z}}_i$ denotes the velocity component of the 
Trend ID at step $i$. By encouraging directional consistency between 
consecutive velocity vectors, $\mathcal{L}_{p}$ suppresses sharp 
turns in the latent trajectory and stabilizes online adaptation.

\section{Temporal Constraints on Trend IDs via a State Transition Model}

\begin{figure}[tb]
    \centering
    \includegraphics[width=\columnwidth]{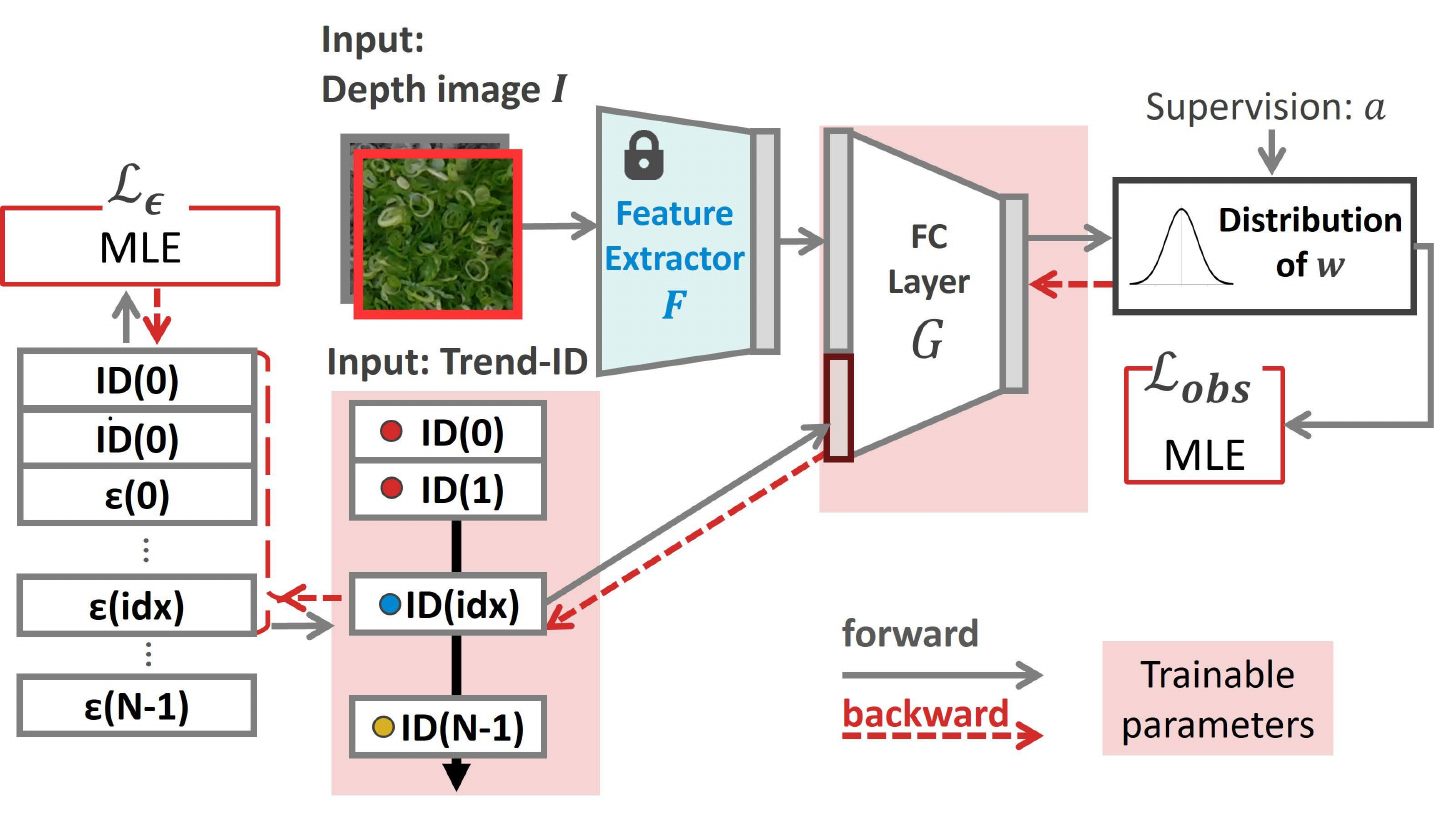}
    \caption{Architecture of the proposed framework. The frozen feature extractor $F$ and trainable FC layer $G$ receive the input and Trend ID to output a predicted distribution. Trend IDs are derived from the state transition model (left), and trainable parameters (shaded) are optimized via backpropagation (dashed arrows).}
    \label{fig:model_v2}
\end{figure}

In the method presented in Section~\ref{sec:method}, an independent Trend ID was assigned to each sample, and temporal continuity was encouraged through pairwise distance penalties. However, this constraint is weak, and the risk remains that each Trend ID overfits independently to the observed data. In this section, under the assumption that the time constant of latent environmental state changes is sufficiently larger than the sampling interval induced by actions, we introduce structural constraints based on a state transition model to impose stronger regularization on the temporal evolution of Trend IDs.

\subsection{General Formulation}
The trend at each time step is represented as a state vector $\mathbf{Z}_i$, and constrained by a state transition model based on the time difference $\Delta t_i = t_i - t_{i-1}$ between adjacent samples, as illustrated in Fig.~\ref{fig:model_v2}:
\begin{equation}
\label{eq:general_transition}
\mathbf{Z}_i = h(\mathbf{Z}_{i-1}, \Delta t_i) + \mathbf{B}\boldsymbol{\epsilon}_i,
\end{equation}
where $h(\cdot)$ is the state transition function and $\boldsymbol{\epsilon}_i$ is the process noise. With this formulation, the learnable quantities become the initial state $\mathbf{Z}_0$ and the process noise sequence $\{\boldsymbol{\epsilon}_i\}_{i=1}^{N-1}$. A loss term that penalizes large process noise effectively limits the degrees of freedom of the model, thereby suppressing overfitting.

The total training loss is given by Eq.~\eqref{eq:total_loss},
where $\mathcal{L}_{\varepsilon}$ is concretized via the
constant-velocity state transition model as described below.

\subsection{Implementation via a Constant-Velocity Motion Model}

In this paper, we adopt a constant-velocity motion model as a concrete instantiation of Eq.~\eqref{eq:general_transition}. This model is grounded in the assumption that the environmental state evolves at an approximately constant rate of change over short time scales. Depending on the characteristics of environmental variation, alternative dynamical systems can be substituted, including models that incorporate acceleration or nonlinear state transitions.

\subsubsection{State-Space Representation and Recurrence Relation}
In the constant-velocity motion model, in addition to the Trend ID position $\mathbf{z}_i \in \mathbb{R}^d$, its rate of change $\dot{\mathbf{z}}_i \in \mathbb{R}^d$ is explicitly modeled, and the state vector is defined as $\mathbf{Z}_i = [\mathbf{z}_i; \dot{\mathbf{z}}_i]$, as shown in Fig.~\ref{fig:model_v2}. Letting $\Delta t_i = t_i - t_{i-1}$ denote the time difference between adjacent samples, the state transition is described by the following recurrence relation:
\begin{equation}
\mathbf{Z}_i = \mathbf{A}(\Delta t_i)\mathbf{Z}_{i-1} + \mathbf{B}\boldsymbol{\epsilon}_i,
\label{eq:cv_transition}
\end{equation}
where the state transition matrix $\mathbf{A}(\Delta t_i)$ encodes the update of position proportional to the rate of change:
\begin{equation}
\mathbf{A}(\Delta t_i) = \begin{bmatrix} \mathbf{I} & \Delta t_i \mathbf{I} \\ \mathbf{0} & \mathbf{I} \end{bmatrix}.
\end{equation}
The process noise $\boldsymbol{\epsilon}_i \sim \mathcal{N}(\mathbf{0}, \sigma_{\epsilon}^2 \Delta t_i \mathbf{I})$ represents stochastic perturbations applied to the rate of change, capturing environmental variations that cannot be predicted by the state transition model alone. By designing the noise variance to scale proportionally with the time difference $\Delta t_i$, the placement of Trend IDs is given greater flexibility at locations with larger time differences, while stronger continuity is enforced where the time difference is small.

\subsubsection{Concrete Form of the Loss Function}
Under the constant-velocity motion model, the state transition loss 
$\mathcal{L}_{\varepsilon}$ is concretized as the negative 
log-likelihood of the process noise $\boldsymbol{\varepsilon}_i$:
\begin{equation}
\mathcal{L}_{\varepsilon} = \sum_{i=1}^{N-1} 
  \frac{\|\boldsymbol{\varepsilon}_i\|_2^2}{\sigma_{\varepsilon}^2 \Delta t_i}.
\label{eq:cv_loss}
\end{equation}
This loss encourages the Trend IDs to follow a smooth trajectory that 
approximates constant-velocity motion by keeping the process noise 
small. Deviations from constant-velocity motion are permitted only 
when required to fit the observed data.

The velocity consistency loss $\mathcal{L}_{v}$ and position 
consistency loss $\mathcal{L}_{p}$ are as defined in 
Eqs.~\eqref{eq:loss_v} and~\eqref{eq:loss_p}, respectively, and 
apply identically under the constant-velocity motion model. The 
hyperparameter $\sigma_{\varepsilon}^2$ reflects the characteristic 
scale of the rate of change of the environmental state. Smaller values 
enforce smoother transitions, while larger values permit the model to 
track relatively rapid environmental changes.

\subsubsection{Test-Time Adaptation}
At test time, the parameters of the feature extractor $F$ and the fully connected layer $G$ are held fixed, as depicted in Fig.~\ref{fig:model_v2}. Given a small number of samples $\{(x_j^{\text{test}}, a_j^{\text{test}}, y_j^{\text{test}})\}_{j=1}^{M}$ collected from a new environment, only the initial state $\mathbf{Z}_0^{\text{test}}$ and the process noise sequence $\{\boldsymbol{\epsilon}_j^{\text{test}}\}_{j=1}^{M-1}$ are optimized. The structural constraints imposed by the state transition model ensure that the estimated Trend IDs are temporally consistent, making it possible to sequentially update the process noise sequence as new observations become available during testing, thereby tracking environmental variation online.

\section{Experiments}

\subsection{Experimental Setup}

\subsubsection{Task Description}

We evaluate the proposed framework on a quantitative grasping task
of granular and fragmented food materials using SCARA-type robots.
The objective is to execute a grasping action based on visual input
such that the resulting grasped weight approaches a target value.

At each trial, the robot observes a candidate grasp region on a food tray
using a depth camera. A grasp parameter, namely the insertion depth,
is determined and executed. The grasped material is then weighed using
an electronic scale to obtain the ground-truth outcome.

This task is inherently subject to concept shift.
Even when visual observations remain similar,
latent environmental factors such as moisture content,
particle size distribution, and packing density
induce substantial variation in the resulting grasped weight.
Accordingly, the problem is formulated as probabilistic regression,
where the model predicts the conditional distribution of the grasped weight
given the observation and action.

The experimental configuration follows the general setup of multi-environment
robotic food handling systems such as \cite{mori2025ros2}.
Note that the SCARA-type robots deployed across the three factories
differ in both the robot arm model and the end-effector design,
introducing additional variability in the mechanical characteristics
of the grasping system beyond the environmental concept shift.

\subsubsection{Dataset}

The dataset was collected from three distinct factories
(Factory A, Factory B, and Factory C)
under different environmental conditions and dates.
Two object types were considered:
chopped green onions and sliced chili peppers.
For each object type, 10 independent time-series sequences were collected,
resulting in a total of 20 sequences.

Each sequence corresponds to an independent experimental session
conducted under a unique environmental condition.
Even within the same factory, sessions recorded on different dates
are treated as distinct environments.
Each sequence consists of 45 temporally ordered samples,
yielding a total of 900 samples across the entire dataset.

Each data sample consists of:
(i) a depth image of the candidate grasp region,
(ii) a randomly assigned insertion depth,
and (iii) the measured grasped weight.

\subsubsection{Train/Test Split}

Among the 20 sequences, 18 were used for training
and the remaining two were reserved as unseen test environments.
The training set consists of 9 sequences per object type.
For evaluation, one unseen sequence for green onions
and one for chili peppers were used exclusively for testing.

During testing, model parameters were fixed.
Only the Trend ID associated with the new environment
was optimized using a small number of labeled samples
to perform few-shot adaptation.

This split enables simultaneous evaluation of generalization
across factories, dates, and object types,
as well as rapid adaptation to previously unseen environments.

\subsubsection{Implementation Details}
The model was implemented using PyTorch.
The feature extractor $F$ adopts MobileNet\cite{howard2017mobilenets} pre-trained on ImageNet,
and the fully connected layer $G$ was trained from scratch.
The Adam optimizer was used with a learning rate of $1.0 \times 10^{-3}$.
The weighting coefficients in the loss function were set to
$\alpha = 1$, $\beta = 100$, $\gamma = 1000$, and $\zeta = 1000$.
Training was conducted for approximately 1000 epochs on an NVIDIA RTX A100 GPU.
These settings were kept identical across all experimental conditions.

\subsection{Trend Space Structure Analysis}

\begin{figure}[t]
    \centering
    \includegraphics[width=\linewidth]{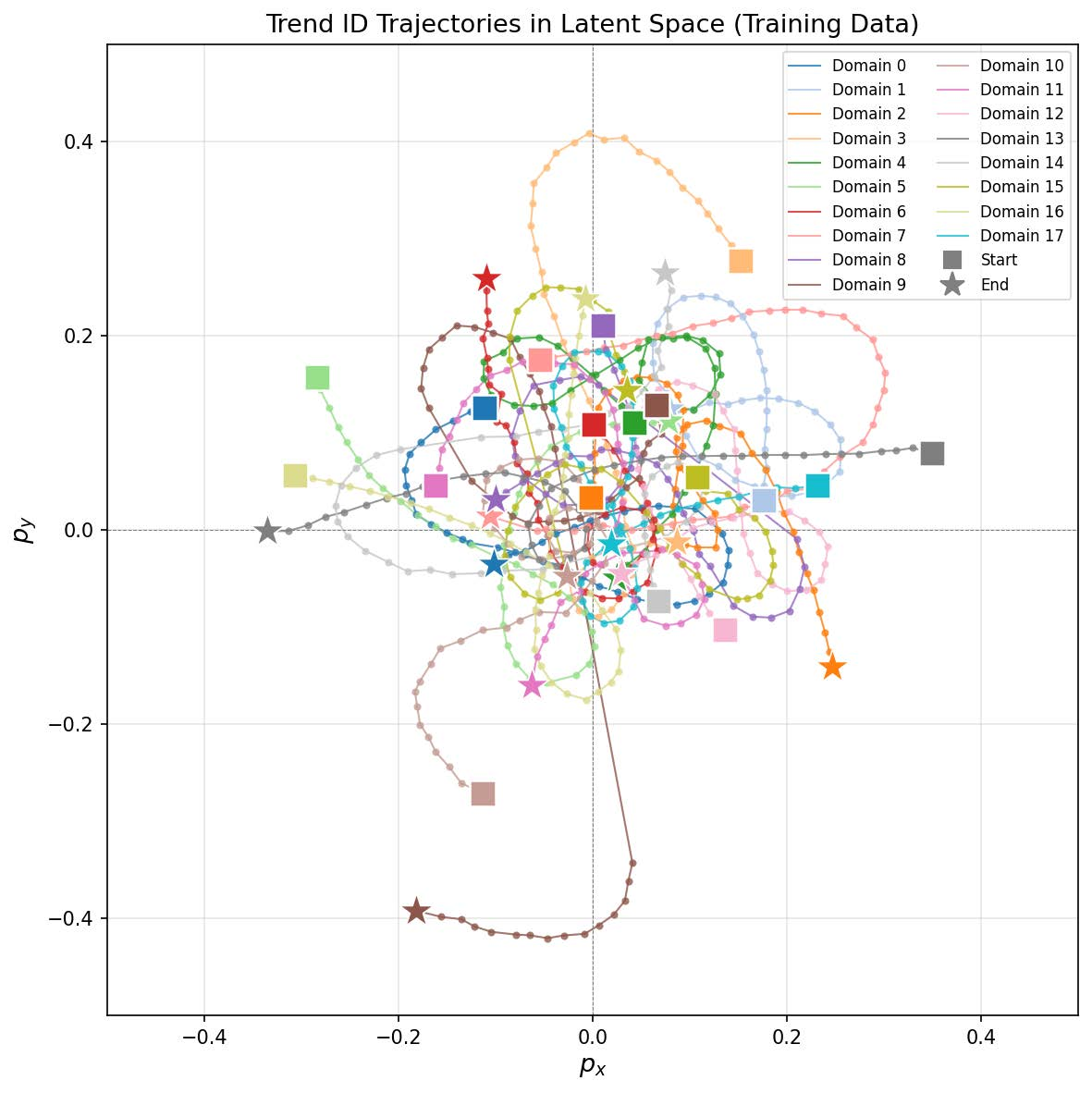}
    \caption{
    Structured trend space constructed during training.
    Each trajectory corresponds to a time sequence collected 
    under a specific factory and date condition.
    Different colors denote different environmental conditions 
    (factory/date/object type).
    The smooth trajectories indicate that the temporal constraint 
    successfully enforces continuity in the latent space.
    }
    \label{fig:trained_space}
\end{figure}

\begin{figure}[t]
    \centering
    \includegraphics[width=\linewidth]{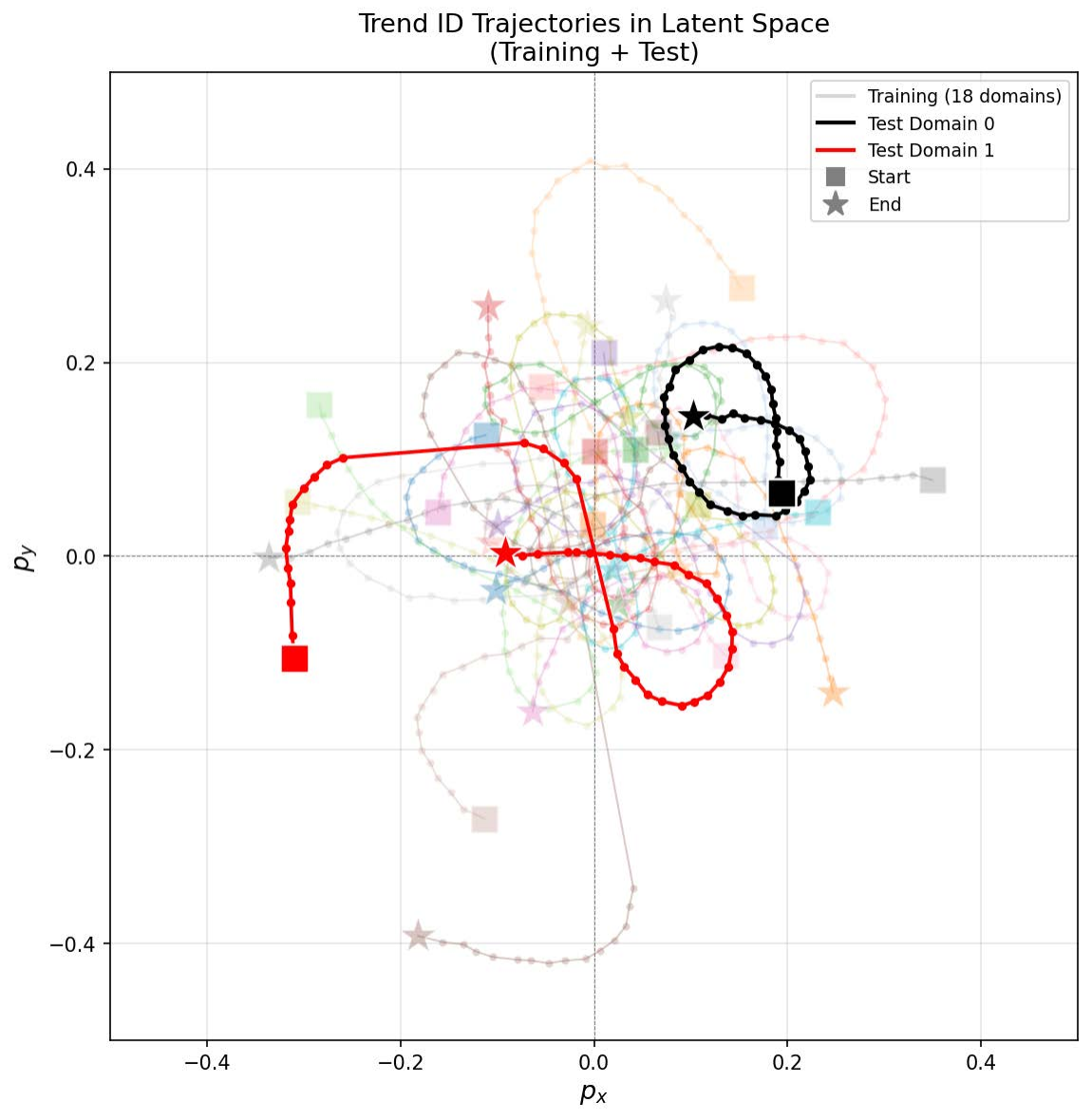}
    \caption{
    Visualization of the estimated Trend IDs in the test
environment after few-shot adaptation. Each point represents
a Trend ID in the 2D latent space ($p_x$, $p_y$).
The estimated Trend IDs are placed within the region spanned
by the training latent space, suggesting that the model
integrates the unseen environment into the existing latent
structure using a small number of samples.
    }
    \label{fig:test_data}
\end{figure}

\subsubsection{Visualization of the Learned Trend Embedding}

Fig.~\ref{fig:trained_space} visualizes the learned Trend IDs
projected onto a two-dimensional latent space.
Each trajectory corresponds to a time-series sequence
collected under a specific factory/date/object condition.

Individual sequences are distributed across distinct regions
of the latent space, indicating that the model embeds different
environmental conditions as separate representations.
Moreover, each sequence exhibits a locally consistent trajectory,
suggesting that the temporal regularization via the state transition
model encourages smooth and coherent evolution of the latent state
within each session.
These observations imply that the Trend ID implicitly captures
environmental variation across sequences, even though an explicit
correspondence to human-interpretable attributes such as factory,
material type, or hand configuration has not been confirmed
under the current hyperparameter settings.

\subsubsection{Few-Shot Adaptation in Unseen Environments}

Fig.~\ref{fig:test_data} shows the Trend IDs estimated after
few-shot optimization in previously unseen test environments.
Despite using only a small number of labeled samples,
the estimated Trend IDs are placed within the region spanned
by the training latent space, without disrupting its global structure.
This indicates that the proposed few-shot adaptation mechanism
successfully integrates new environmental states into the existing
latent space while keeping all model parameters fixed,
thereby preserving previously acquired knowledge.

\subsection{Discussion}

The experimental results provide evidence for two key properties
of the proposed framework.

First, the Trend ID embedding distributes different environmental
sessions across distinct regions of the latent space,
and each session exhibits temporally consistent trajectories.
This suggests that the framework implicitly responds to concept shift
by assigning each environment a unique and coherent latent
representation, even in the absence of explicit supervision
on environmental attributes.

Second, few-shot adaptation to unseen environments is achieved
without modifying model parameters.
The test Trend IDs converge to locations consistent with the
training latent space using only a small number of observations,
demonstrating that the proposed method avoids catastrophic
forgetting by construction.

It should be noted that, under the current experimental settings,
a clear geometric correspondence between the latent space structure
and human-interpretable attributes---such as factory identity,
material type, or hand configuration---was not conclusively observed.
This is likely attributable to the sensitivity of the latent space
organization to the choice of loss weighting coefficients
$(\alpha, \beta, \gamma, \zeta)$ and the regularization design,
rather than a fundamental limitation of the proposed framework.
Systematic exploration of these hyperparameters, or the incorporation
of more expressive state transition models, is expected to yield
a more interpretable latent structure in future work.

\section{Conclusion}

This paper proposed a concept shift adaptation framework
based on latent Trend ID embedding for robotic systems
operating in non-stationary environments.
Unlike conventional approaches that update model parameters
during adaptation, the proposed method maintains fixed model weights
and instead adapts a low-dimensional latent environmental state,
referred to as the Trend ID.
This formulation enables rapid few-shot adaptation
while inherently avoiding catastrophic forgetting.

To prevent overfitting caused by assigning learnable IDs to each sample,
we introduced structured regularization through temporal constraints
and further strengthened this structure using a state transition model.
In particular, a constant-velocity motion model was incorporated
to enforce smooth temporal evolution of the latent environmental state,
thereby reducing the degrees of freedom of the adaptation process
while preserving responsiveness to environmental changes.

Experimental validation was conducted on a quantitative
grasping task of granular food materials, where unobservable
environmental factors such as moisture and density induce
significant concept shift. The results demonstrated that: (1)
different environmental sessions are embedded as distinct
representations in the latent space with temporally coherent
trajectories, suggesting implicit adaptation to concept shift,
(2) temporal environmental drift is represented as smooth
trajectories enforced by the state transition model, and (3)
rapid few-shot adaptation to unseen environments is achieved
without updating model parameters.

These findings indicate that the proposed framework
provides a scalable and interpretable solution
for robotics applications in multi-environment settings,
such as multi-site production lines and long-term deployment scenarios,
where environmental conditions evolve continuously and recur over time.

Future work includes extending the state transition model
to more expressive nonlinear dynamical systems,
integrating online uncertainty estimation of the latent state,
and applying the framework to broader robotic tasks
including manipulation, locomotion, and multi-robot coordination.

\section*{Acknowledgement}
This work was supported by JSPS KAKENHI Grant Number JP 24K03021and NIPPN CORPORATION.

\bibliographystyle{IEEEtran}
\bibliography{refs}

\end{document}